\documentclass{article}

\usepackage{xcolor}
\usepackage[preprint]{corl_2026} 

\usepackage{graphicx}
\usepackage{amsmath}
\usepackage{booktabs}
\usepackage{multirow}
\usepackage{wrapfig}
\usepackage{makecell}
\usepackage[table]{xcolor}
\usepackage{enumitem}
\usepackage{amssymb}
\usepackage{titletoc}

\title{Ego2Robot: Scalable Robot Data Synthesis from Egocentric Human Data}

\author{
  \textbf{Ye Wang\textsuperscript{1,2,$*$},
  Pei Lin\textsuperscript{2,3,4,$*$},
  Xiong-Hui Chen\textsuperscript{2,$*$},
  Haoqi Yuan\textsuperscript{2},
  Zhixuan Liang\textsuperscript{2},} \\
  \textbf{Yiyang Huang\textsuperscript{2},
  Anzhe Chen\textsuperscript{2}, 
  Zixing Lei\textsuperscript{2},
  Jie Zhang\textsuperscript{2},
  Tao Zhang\textsuperscript{1},
  Haoyang Li\textsuperscript{2},} \\
  \textbf{Tong Zhang\textsuperscript{2},
  Chenxi Xiao\textsuperscript{3},
  Ziyuan Jiao\textsuperscript{4,5},
  Qin Jin\textsuperscript{1,$\dagger$}} \\
  \textsuperscript{1}AIM3 Lab, Renmin University of China \quad
  \textsuperscript{2}Qwen Team, Alibaba Inc. \\
  \textsuperscript{3}ShanghaiTech University \quad
  \textsuperscript{4}Beijing Institute for General Artificial Intelligence (BIGAI) \\ \textsuperscript{5}Beijing University of Aeronautics and Astronautics  \\
  \small{\textsuperscript{$*$}Equal contribution \quad \textsuperscript{$\dagger$}Corresponding author}
}

\begin{document}
\maketitle


\vspace{-15pt}

\begin{abstract}
Learning generalizable robot manipulation policies requires large-scale and diverse demonstration data. 
Egocentric human manipulation videos offer rich scene and task diversity, and prior work has shown that retargeting and rendering such videos into robot-format data can yield effective per-task policies at small scale. 
However, whether this approach can provide pretraining benefits for vision-language-action models at scale remains unexplored.
We present \textbf{Ego2Robot}, a scalable pipeline that converts egocentric human manipulation videos into robot training data through action retargeting, robot-arm visual synthesis, and multi-level quality curation. 
Ego2Robot supports both curated datasets and in-the-wild videos, producing 18,561 hours of robot training data spanning 15 robot morphologies, making it the largest ego-to-robot dataset to date. 
To evaluate generalization, we extend RoboTwin2.0 with disentangled perturbation axes covering visual appearance, scene layout, embodiment morphology, and task semantics. 
Experiments show that joint pretraining on Ego2Robot-synthesized and robot data consistently improves out-of-distribution generalization across multiple perturbation types, with benefits validated on real-robot deployment.
Project page: \url{https://www-ye.github.io/ego2robot_blog/}

\end{abstract}



\section{Introduction}
\label{sec:intro}

\begin{figure*}[htbp]
    \centering
    \includegraphics[width=\textwidth]{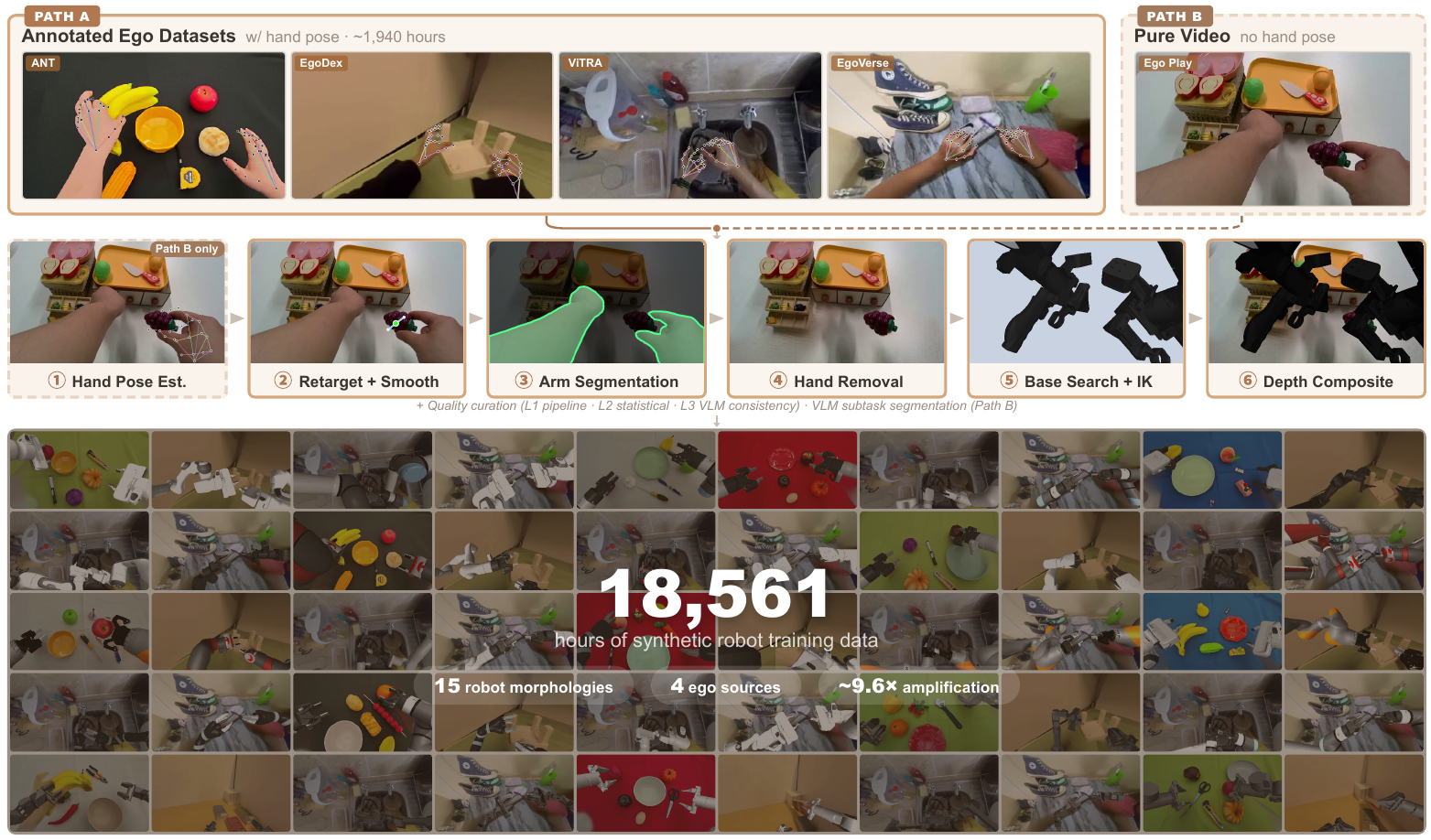}
    \vspace{-15pt}
    \caption{\textbf{Ego2Robot pipeline.} Converting egocentric video into 18{,}561h robot training data across 15 morphologies through action alignment, visual alignment, and quality curation.}
    \vspace{-15pt}
    \label{fig:pipeline}
\end{figure*}

Recent vision-language-action (VLA) models~\cite{zitkovich2023rt,kim2024openvla,black2024pi_0,intelligence2025pi05,bjorck2025gr00t} have demonstrated impressive progress in robot manipulation through large-scale robot demonstration pretraining. 
However, the generalization ability of these systems remains fundamentally constrained by the scale and diversity of available robot data. 
Despite major efforts such as Open X-Embodiment~\cite{o2024open}, DROID~\cite{khazatsky2024droid}, and AgibotWorld~\cite{bu2025agibot}, collecting robot demonstrations remains expensive, labor-intensive, and limited by hardware availability, teleoperation cost, and restricted interaction diversity~\cite{chi2024universal,zhao2023learning,wu2024gello}.

Egocentric human videos offer a compelling alternative. 
Compared with robot teleoperation, human hand interactions can be collected at massive scale across diverse objects, environments, and task variations~\cite{hoque2025egodex,punamiya2026egoverse}. 
These videos capture rich manipulation priors that are difficult to obtain with robots alone. 
However, directly converting egocentric videos into robot training data remains challenging due to the substantial gap between human and robot embodiments. 
One promising direction is to retarget hand motions to robot kinematics and visually replace human arms with rendered robot embodiments, which has shown strong per-task results at small scale~\cite{lepert2025phantom,kareer2025egomimic}. 
Whether large-scale ego2robot-synthesized data can serve as effective pretraining data for robot policies, particularly for improving out-of-distribution generalization, remains largely unexplored.

In this work, we investigate ego-to-robot synthesis as a scalable paradigm for robot policy pretraining. 
Our key hypothesis is that, despite embodiment differences, egocentric manipulation videos contain transferable interaction regularities that can complement robot data if properly aligned. 
Based on this observation, we propose \textbf{Ego2Robot}, an end-to-end pipeline that converts egocentric human manipulation data into embodiment-specific robot training data through action alignment, visual alignment, and multi-level quality curation (Figure~\ref{fig:pipeline}).
The pipeline supports both annotated ego datasets and pure ego videos, and processes approximately 1,940 hours of egocentric video from four diverse sources across 15 robot morphologies, producing 18,561 hours of effective robot training data, to the best of our knowledge the largest ego-to-robot dataset to date.

To evaluate whether ego2robot-synthesized data improves robot generalization, we further introduce a disentangled evaluation protocol built on RoboTwin2.0~\cite{chen2025robotwin}. 
Unlike prior evaluations that aggregate multiple distribution shifts into a single score, our protocol decouples perturbations across four axes: visual appearance, scene layout, embodiment morphology, and task semantics. 
This enables fine-grained analysis of where pretraining with a combination of robot and ego2robot-synthesized data provides benefits and whether the gains arise from visual robustness, embodiment transfer, or semantic generalization.
Extensive experiments show that ego2robot-synthesized data consistently improves out-of-distribution performance and provides complementary value to robot data. 
In particular, the gains are most pronounced under visual, embodiment, and semantic perturbations, suggesting that large-scale ego2robot-synthesized data primarily improves invariance and cross-distribution robustness rather than merely increasing trajectory coverage.

Our contributions are as follows:
(1) We propose Ego2Robot, a complete ego-to-robot synthesis pipeline with action alignment, visual alignment, and multi-level quality curation, supporting 15 robot morphologies. 
(2) We construct the largest ego-to-robot dataset to date, containing 18,561 hours of synthesized robot training data generated from diverse egocentric video sources. 
(3) We introduce a disentangled generalization benchmark that separates visual, scene, embodiment, and semantic perturbations for fine-grained evaluation of robot policy robustness. 
(4) We demonstrate that combining ego2robot-synthesized data with pure robot data improves generalization performance, particularly under out-of-distribution shifts.

\section{Related Work}
\label{sec:related}

\paragraph{Robot Data Scaling and Learning from Human Data}

Scaling robot demonstration data is central to generalizable manipulation. 
Large-scale robot datasets~\cite{o2024open,khazatsky2024droid,bu2025agibot} and portable collection systems~\cite{zhao2023learning,wu2024gello,chi2024universal,shafiullah2023bringing,DexMove_ICLR2026} have expanded available demonstrations, but diversity remains constrained by collection cost and hardware scalability. 
Human manipulation videos provide a broader alternative, motivating work on visual pretraining~\cite{nair2022r3m,radosavovic2023real,ma2022vip}, reward learning~\cite{chen2021learning}, motion priors~\cite{wang2023mimicplay}, and point tracking~\cite{xu2024flow,ren2025motion}, though these methods still rely on robot data to bridge embodiment gaps.

More direct approaches leverage human data for robot policy learning. 
One line of work co-trains on egocentric or human video alongside robot data~\cite{kareer2025egomimic,zheng2026egoscale,luo2025being,luo2026being,li2025scalable}, while another converts human demonstrations into robot-format data through retargeting and rendering~\cite{lepert2025phantom,zhang2026easymimic} or embodiment masking~\cite{kareer2025egomimic}. 
However, the retarget-and-render approach has only been studied at limited scale or for individual tasks. 
Our work scales it to 18{,}561 hours across 15 robot morphologies and systematically evaluates its benefit for VLA pretraining and out-of-distribution generalization.

\paragraph{Data Augmentation for Robot Learning}
Data augmentation expands training diversity without additional robot collection. 
Simulation-based methods~\cite{mandlekar2023mimicgen,nasiriany2024robocasa} generate synthetic demonstrations in digital twins, while cross-embodiment methods including RoviAug~\cite{chen2024rovi}, Mirage~\cite{chen2024mirage} and OXE-AugE~\cite{ji2025oxe} bridge robot-to-robot visual gaps through inpainting and image editing. 
Unlike these approaches, our work extends augmentation to the substantially larger \emph{ego-to-robot} domain gap, leveraging egocentric human videos to synthesize training data for 15 robot morphologies and systematically evaluating generalization across multiple perturbation axes.

\paragraph{Benchmarks for Robot Generalization}

Existing manipulation benchmarks such as RoboTwin~\cite{mu2025robotwin,chen2025robotwin}, LIBERO~\cite{liu2023libero}, CALVIN~\cite{mees2022calvin}, RLBench~\cite{james2020rlbench}, and SIMPLER~\cite{li2024evaluating} evaluate policy generalization under bundled perturbations, making it difficult to attribute failures to specific factors. ManiSkill2/3~\cite{gu2023maniskill2,tao2024maniskill3} and RoboCasa~\cite{nasiriany2024robocasa} support multiple robot configurations but lack standardized cross-embodiment evaluation. 
Recent decomposed benchmarks, including Colosseum~\cite{pumacay2024colosseum}, LIBERO-Plus~\cite{fei2025liberoplus}, and LIBERO-PRO~\cite{zhou2025liberopro}, separate perturbation axes and show that VLA models degrade substantially under individual shifts, but remain limited to single-arm settings. 
RoboTwin 2.0~\cite{chen2025robotwin} and EBench~\cite{ebench2026} evaluate dual-arm manipulation yet still primarily report bundled results. 
We extend RoboTwin2.0 with independent perturbation axes spanning visual appearance, scene layout, embodiment, and task semantics, while jointly supporting dual-arm and cross-embodiment evaluation.

\section{Ego2Robot Pipeline}
\label{sec:pipeline}

Given egocentric videos depicting human hand manipulation, our pipeline produces embodiment-specific robot training data through three stages: \emph{action alignment}, \emph{visual alignment}, and \emph{quality curation} (Figure~\ref{fig:pipeline}). 
Action alignment converts hand poses into robot end-effector trajectories via retargeting and temporal smoothing.
Visual alignment replaces human arms with rendered robot arms through arm segmentation, hand removal, robot base pose search with IK solving, and depth-aware rendering. 
Quality curation then filters samples at the trajectory, frame, and episode levels.
The pipeline supports two input paths: \textbf{Path~A} accepts ego datasets with existing hand pose annotations, while \textbf{Path~B} processes unannotated videos by first estimating hand poses via WiLoR~\cite{potamias2025wilor,si2026anyhand} per-frame reconstruction and DynHaMR~\cite{yu2025dyn} temporal optimization. 
For long videos, we adopt Qwen3.5~\cite{qwen35blog} to segment continuous recordings into discrete subtasks with natural language descriptions. 
After hand pose estimation, both paths share a unified pipeline that generates training data for any of 15 supported robot morphologies in parallel.

\paragraph{1. Action Alignment}
\label{sec:action}

The action alignment stage converts hand poses into parallel-gripper end-effector trajectories through retargeting and temporal smoothing.

\textit{\textbf{Hand-to-Gripper Retargeting.}}
For each frame with a detected hand, we extract a compact gripper representation from the 21 hand keypoints. 
We define a virtual fingertip as a weighted blend of the index and middle finger tips:
\begin{equation}
\small
    \mathbf{p}_\text{vf} = 0.7 \cdot \mathbf{p}_\text{index} + 0.3 \cdot \mathbf{p}_\text{middle}
\end{equation}
The tool center point (TCP) and gripper opening width are:
\begin{equation}
\small
    \mathbf{p}_\text{tcp} = \frac{\mathbf{p}_\text{thumb} + \mathbf{p}_\text{vf}}{2}, \quad
    w = \|\mathbf{p}_\text{thumb} - \mathbf{p}_\text{vf}\|
\end{equation}
The grasp orientation is a right-handed orthonormal frame $\mathbf{R} = [\mathbf{x}\ \mathbf{y}\ \mathbf{z}]$.
The grasp axis $\mathbf{z}$ lies along the jaw line (thumb tip to virtual fingertip);
together with the wrist-to-fingertip direction $\mathbf{d} = \mathbf{p}_\text{vf} - \mathbf{p}_\text{wrist}$
it spans the jaw plane, whose normal is the gripper-normal axis $\mathbf{y}$; the approach axis $\mathbf{x}$ completes the frame:
\vspace{-5pt}\begin{equation}
\small
    \mathbf{z} = \frac{s\,(\mathbf{p}_\text{thumb} - \mathbf{p}_\text{vf})}{w}, \quad
    \mathbf{y} = \frac{\mathbf{z} \times \mathbf{d}}{\|\mathbf{z} \times \mathbf{d}\|}, \quad
    \mathbf{x} = \mathbf{y} \times \mathbf{z}
\end{equation}
where $s = +1$ for the right hand and $s = -1$ for the left hand, so that $\mathbf{z}$ points consistently
regardless of handedness and both hands map to the same gripper frame.
The three axes are: $\mathbf{x}$ – approach direction, $\mathbf{y}$ – gripper normal (perpendicular to the jaw plane),
$\mathbf{z}$ – grasp axis (along the jaw line).

\textit{\textbf{Temporal Smoothing.}}
Per-frame hand detection introduces high-frequency noise. 
We apply Savitzky-Golay filtering to positions and widths, and Gaussian-weighted SLERP to orientations, producing smooth trajectories while preserving motion structure.

\textit{\textbf{Action Speed Alignment.}}
Egocentric hand manipulation exhibits significantly higher action speeds than robot teleoperation data.
To align the speed distributions, we apply per-source frame subsampling during training:
ANT and EgoDex are downsampled to 60\% of their original frame rate ($\sim$1.7$\times$ slower),
EgoVerse to 45\% ($\sim$2.2$\times$ slower), and ViTRA to 25\% ($\sim$4$\times$ slower).

\paragraph{2. Visual Alignment}
\label{sec:visual}

Visual alignment transforms ego video from depicting human hands to showing a robot arm operating in the same scene.

\textit{\textbf{Arm Segmentation.}}
We employ SAM~3~\cite{carion2025sam3} to segment human arm regions in each frame, providing temporally consistent masks across frames.

\textit{\textbf{Hand Removal.}}
With the arm masks, ProPainter~\cite{zhou2023propainter} performs temporally consistent video inpainting to remove the human arms and reconstruct the background.

\textit{\textbf{Robot Base Pose Search.}}
A fundamental challenge in converting ego videos to robot data is determining the robot base placement. 
Unlike robot-to-robot transfer where a source base position is available, egocentric hand trajectories are embodiment-free: there is no physical robot base to reference. 
We must find a base pose $\mathbf{T}_\text{base} = (\mathbf{t}, \mathbf{R}) \in SE(3)$ such that the retargeted trajectory remains kinematically feasible for the target robot morphology.

We formulate this as an optimization over base placements. 
Given a trajectory of $N$ target end-effector poses $\{\mathbf{T}_i^\text{ee}\}_{i=1}^{N}$ and a robot with maximum reach $r_\text{max}$, we seek:
\begin{equation}
\small
    \mathbf{T}_\text{base}^* = \arg\max_{\mathbf{T}_\text{base}} \frac{1}{|\mathcal{K}|} \sum_{k \in \mathcal{K}} \mathbb{1}\left[\text{IK}(\mathbf{T}_\text{base}^{-1} \mathbf{T}_k^\text{ee}) \text{ is feasible}\right]
\end{equation}
where $\mathcal{K} \subset \{1, \dots, N\}$ is a set of representative keyframes selected to cover the spatial extremes of the trajectory (positions with maximum displacement or orientation change), and $\text{IK}(\cdot)$ denotes an inverse kinematics solver in MuJoCo~\cite{todorov2012mujoco,Zakka_Mink_Python_inverse_2026}. 
Candidate base placements are generated via grid search around the trajectory centroid, constrained by the per-morphology kinematic reach $r_\text{max}$. 
Each candidate is validated by solving IK at all keyframes, and the placement with the highest feasibility rate is selected.
This search is performed independently for each of the 15 robot morphologies, as different arm lengths and joint configurations require different base placements for the same trajectory.

\textit{\textbf{Inverse Kinematics and Rendering.}}
Given the optimized base placement, IK is solved frame-by-frame in MuJoCo. 
The robot is rendered from the original camera viewpoint.

\textit{\textbf{Depth-Aware Compositing.}}
The robot is rendered from the camera viewpoint and composited into the inpainted scene using depth ordering:
\begin{equation}
\small
    I_\text{final}(u,v) = \begin{cases}
        I_\text{robot}(u,v) & \text{if } D_\text{robot}(u,v) < D_\text{scene}(u,v) \land M_\text{robot}(u,v) = 1 \\
        I_\text{inpaint}(u,v) & \text{otherwise}
    \end{cases}
\end{equation}
where $D_\text{scene}$ is obtained from depth sensors or estimated via monocular depth. This process is applied independently for each of 15 robot morphologies (Panda, UR5e, ARX-L5, xArm7, Sawyer, Kinova~Gen3, IIWA, Jaco, FR3, UR10e, ViperX, WidowX, Piper, YAM, Aloha-Agilex), generating parallel training streams from each ego video.

\paragraph{3. Quality Curation}
\label{sec:curation}

We apply three-level quality filtering. 
\textbf{L1 (Pipeline-internal):} frames with IK failures, self-collisions, action outliers, or insufficient workspace coverage are flagged during processing. 
\textbf{L2 (Statistical):} trajectories with extreme action values, sudden discontinuities, or excessive invalid frame ratios are removed. 
\textbf{L3 (VLM Consistency):} a vision-language model~\cite{qwen35blog} audits synthesized videos for semantic consistency between rendered robot actions and original manipulation intents.


We apply the pipeline to four egocentric sources: ANT (7h, our in-house pick-and-place dataset with hand pose annotations), EgoDex~\cite{hoque2025egodex} (732h), ViTRA~\cite{li2025scalable} (249h), and EgoVerse~\cite{punamiya2026egoverse} (954h), totaling $\sim$1,940 hours of annotated ego data. 
Because egocentric hand manipulation moves considerably faster than robot teleoperation,
we subsample frames per source during training to match the robot speed distribution:
ANT and EgoDex to 60\% of their original frame rate ($\sim$1.7$\times$ slower),
EgoVerse to 45\% ($\sim$2.2$\times$ slower), and ViTRA to 25\% ($\sim$4$\times$ slower).

After processing across 15 robot morphologies and quality curation, this yields 18,561 hours of synthetic robot training data. 
Each sample consists of a robot-composited video frame, the corresponding camera-frame EEF action, camera parameters, and a text instruction. 
Because egocentric videos are captured with diverse, unknown camera placements, a world-frame action representation would require per-video calibration and produce incompatible action spaces across sources. 
Camera-frame relative EEF actions express end-effector displacements in the observer's coordinate frame, naturally unifying data from different camera setups and robot morphologies.

\section{Evaluation Framework}
\label{sec:benchmark}

\begin{figure*}[t]
    \centering
    \includegraphics[width=\textwidth]{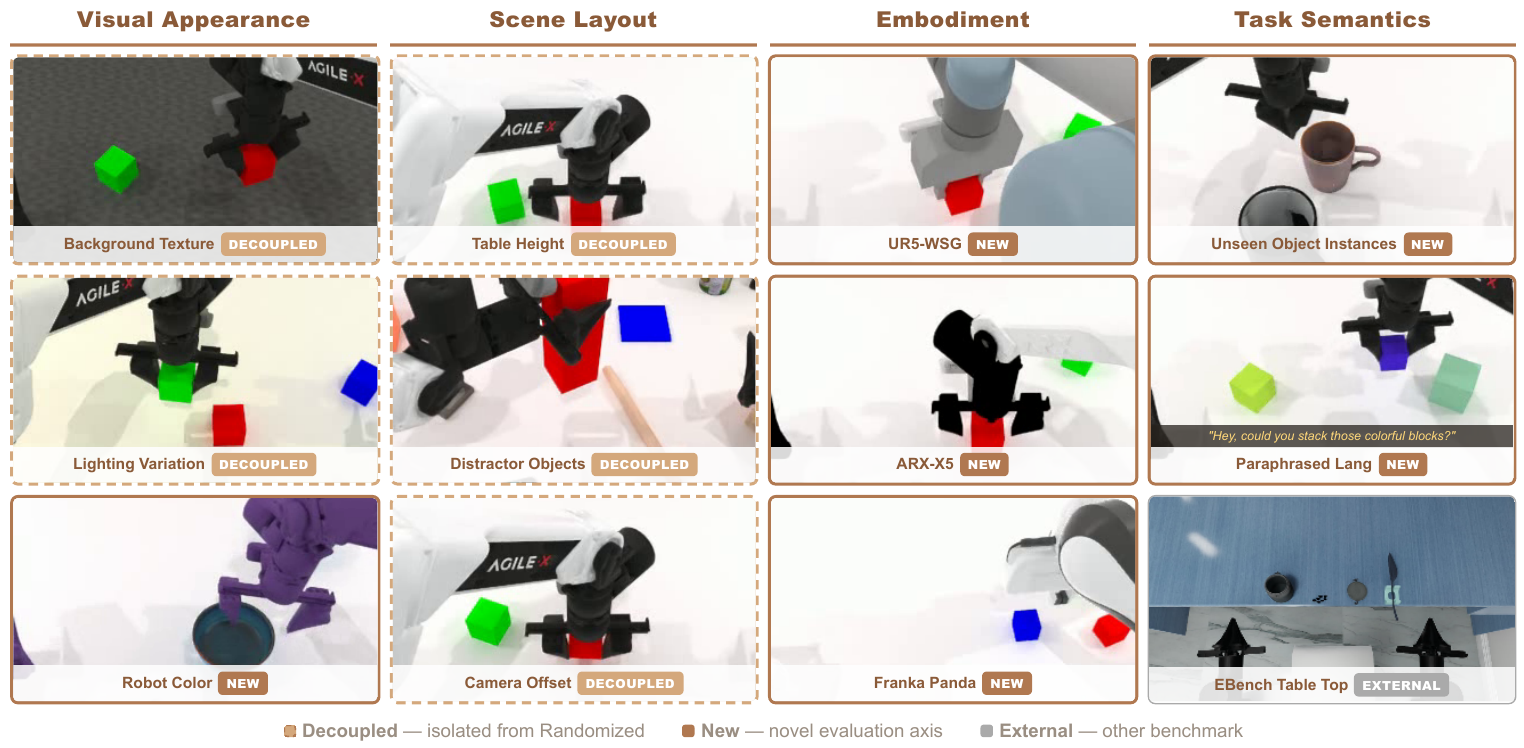}
    \vspace{-15pt}
    \caption{\textbf{Evaluation framework.} 
    Four generalization dimensions with 12 evaluation settings. 
    Dashed borders: settings decoupled from bundled randomization for independent testing.
    Solid borders: newly introduced evaluation axes. 
    Gray border: external benchmark (EBench).}
    \label{fig:benchmark}
    \vspace{-8pt}
\end{figure*}

Understanding which generalization dimensions benefit from ego2robot-synthesized data---and which are limited by domain gaps---requires evaluation granularity that existing benchmarks do not provide. 
Current protocols such as RoboTwin 2.0~\cite{chen2025robotwin} apply multiple perturbation factors simultaneously, producing a single aggregate OOD metric that conflates visual, spatial, embodiment, and semantic shifts.

We extend RoboTwin 2.0 with 11 independent perturbation settings and camera-frame relative EEF support, and complement it with EBench~\cite{ebench2026}, which provides 7 precision tabletop tasks in Isaac Sim with a higher-mounted head camera closer to the egocentric perspective. 
The extension covers four generalization dimensions as shown in Figure~\ref{fig:benchmark}:

(1)~\textbf{Visual Appearance}---independent control of background texture, lighting, and robot color. 
Background and lighting are decoupled from the original bundled randomization for isolated testing. 
Robot color applies random hue shifts to all robot links, testing visual invariance to embodiment appearance changes not seen during training.
(2)~\textbf{Scene Layout}---table height variation ($\pm$4cm), distractor objects, and camera viewpoint offset ($\pm$5cm), all decoupled for independent evaluation. 
These spatial perturbations probe robustness to physical arrangement changes common in real-world deployment.
(3)~\textbf{Embodiment}---replacing the default Aloha-Agilex with UR5-WSG, ARX-X5, and Franka Panda for zero-shot cross-embodiment evaluation, with initial end-effector poses aligned via IK to ensure consistent starting conditions. 
This tests whether the camera-frame action representation generalizes across morphologically distinct robots without embodiment-specific fine-tuning.
(4)~\textbf{Task Semantics}---50 tasks evaluated with object instances unseen during that task's training, and 505 paraphrased natural language instructions that rephrase commands in colloquial form. 
This probes generalization to novel object appearances and robustness to linguistic variation.
This decomposition enables precise attribution of generalization gains to specific perturbation types. 
Further details are provided in the Appendix.

\section{Experiments}
\label{sec:experiments}

\subsection{Experimental Setup}

\textbf{Model Architecture.}
We adopt a VLA architecture with Qwen3.5-4B as the vision-language backbone and a Diffusion Transformer (DiT) action head. 
The model predicts 32-step action chunks in camera-frame relative end-effector representation with 8 diffusion steps. 

\textbf{Training Protocol.}
All pretraining runs use identical hyperparameters: 8 GPUs, batch size 12/GPU, learning rate $1\text{e-5}\rightarrow1\text{e-6}$ with cosine decay, bf16, 200K steps. 
Since all configurations train for the same number of steps with the same batch size, each setting processes an identical number of training samples ($\sim$19.2M frames), ensuring fair comparison regardless of dataset size. Finetuning loads pretrained weights with ColorJitter augmentation for 50K steps.

\textbf{Pretraining Data.}
We compare four configurations: (i) robot-only (DROID~\cite{khazatsky2024droid} + AgibotWorld~\cite{bu2025agibot} + InternData~\cite{tian2025interndata}, $\sim$6,565h), and (ii--iv) Ego2Robot-synthesized data (abbreviated \textbf{Ego2R}) mixed with robot data at 1:3, 3:1, and 1:1 ratios.
\textbf{Finetuning and Evaluation.}
All models are finetuned on RoboTwin's 50 clean-setting tasks (50 demonstrations per task) and evaluated under each perturbation setting with 50 episodes per task. 
EBench models are finetuned separately on its training set. 
For real-robot evaluation, we test on an ARX ACone platform across 5 long-horizon manipulation tasks.
More details are provided in the Appendix.

\subsection{Main Results}

\begin{table*}[t]
\centering
\small
\caption{\textbf{Main results.} Success rates (\%) on RoboTwin2.0 and EBench. \colorbox{green!12}{Green} = gain $>$5\% vs.\ Robot-only.}
\label{tab:main}
\vspace{3pt}
\newcommand{\upa}[1]{{\scriptsize\color{green!50!black}+#1}}
\newcommand{\dna}[1]{{\scriptsize\color{red!70!black}--#1}}
\newcommand{\UPa}[1]{\colorbox{green!12}{\scriptsize\color{green!30!black}\textbf{+#1}}}
\renewcommand{\arraystretch}{1.}
\resizebox{0.8\linewidth}{!}{
\begin{tabular}{l cc cccc c}
\toprule
\multirow{2}{*}{Pretraining} & \multicolumn{2}{c}{RoboTwin} & \multicolumn{4}{c}{Per-Dimension (RoboTwin)} & EBench \\
\cmidrule(lr){2-3} \cmidrule(lr){4-7} \cmidrule(lr){8-8}
 & Clean & Rand & Visual & Scene & Embody & Task & Avg \\
\midrule
Robot-only & 62.2 & 50.9 & 61.4 & 52.9 & 23.8 & 46.2 & 39.6 \\[2pt]
Ego2R+Robot (1:3) & \makecell{61.4\\\dna{0.8}} & \makecell{51.0\\\upa{0.1}} & \makecell{61.2\\\dna{0.2}} & \makecell{52.5\\\dna{0.4}} & \makecell{21.9\\\dna{1.9}} & \makecell{49.5\\\upa{3.3}} & \makecell{47.4\\\UPa{7.8}} \\[2pt]
Ego2R+Robot (3:1) & \makecell{64.1\\\upa{1.9}} & \makecell{49.2\\\dna{1.7}} & \makecell{62.7\\\upa{1.3}} & \makecell{54.3\\\upa{1.4}} & \makecell{\textbf{28.2}\\\upa{4.4}} & \makecell{51.6\\\UPa{5.4}} & \makecell{\textbf{51.7}\\\UPa{12.1}} \\[2pt]
Ego2R+Robot (1:1) & \makecell{\textbf{68.1}\\\UPa{5.9}} & \makecell{\textbf{53.5}\\\upa{2.6}} & \makecell{\textbf{67.3}\\\UPa{5.9}} & \makecell{\textbf{56.9}\\\upa{4.0}} & \makecell{27.2\\\upa{3.4}} & \makecell{\textbf{54.1}\\\UPa{7.9}} & \makecell{49.8\\\UPa{10.2}} \\
\bottomrule
\end{tabular}}
\vspace{-10pt}
\end{table*}

\begin{table*}[t]
\centering
\caption{\textbf{Per-perturbation breakdown.} Change vs.\ Robot-only. \colorbox{green!12}{Green} = gain $>$5\%.}
\label{tab:breakdown}
\setlength{\tabcolsep}{3pt}
\small
\renewcommand{\arraystretch}{1.}
\newcommand{\up}[1]{{\scriptsize\color{green!50!black}+#1}}
\newcommand{\dn}[1]{{\scriptsize\color{red!70!black}--#1}}
\newcommand{\UP}[1]{\colorbox{green!12}{\scriptsize\color{green!30!black}\textbf{+#1}}}
\resizebox{0.95\linewidth}{!}{
\begin{tabular}{l ccc ccc ccc cc}
\toprule
\multirow{2}{*}{Pretraining} & \multicolumn{3}{c}{Visual} & \multicolumn{3}{c}{Scene} & \multicolumn{3}{c}{Embodiment} & \multicolumn{2}{c}{Task} \\
\cmidrule(lr){2-4} \cmidrule(lr){5-7} \cmidrule(lr){8-10} \cmidrule(lr){11-12}
 & BG & Light & Color & Height & Clutter & Camera & ARX & UR5 & Franka & Obj & Lang \\
\midrule
Robot-only & 66.6 & 58.2 & 59.4 & 60.1 & 48.3 & 50.4 & 44.1 & 20.2 & 7.0 & 29.3 & 63.1 \\[2pt]
\makecell[l]{Ego2R+Robot (1:3)} & \makecell{65.0\\\dn{1.6}} & \makecell{58.3\\\up{0.1}} & \makecell{60.3\\\up{0.9}} & \makecell{58.6\\\dn{1.5}} & \makecell{49.3\\\up{1.0}} & \makecell{49.6\\\dn{0.8}} & \makecell{43.7\\\dn{0.4}} & \makecell{17.6\\\dn{2.6}} & \makecell{4.5\\\dn{2.5}} & \makecell{36.8\\\UP{7.5}} & \makecell{62.2\\\dn{0.9}} \\[2pt]
\makecell[l]{Ego2R+Robot (3:1)} & \makecell{65.5\\\dn{1.1}} & \makecell{60.9\\\up{2.7}} & \makecell{61.8\\\up{2.4}} & \makecell{62.0\\\up{1.9}} & \makecell{49.2\\\up{0.9}} & \makecell{51.6\\\up{1.2}} & \makecell{47.6\\\up{3.5}} & \makecell{\textbf{31.4}\\\UP{11.2}} & \makecell{5.6\\\dn{1.4}} & \makecell{\textbf{40.0}\\\UP{10.7}} & \makecell{63.1\\\up{0.0}} \\[2pt]
\makecell[l]{Ego2R+Robot (1:1)} & \makecell{\textbf{70.3}\\\up{3.7}} & \makecell{\textbf{65.8}\\\UP{7.6}} & \makecell{\textbf{65.8}\\\UP{6.4}} & \makecell{\textbf{62.4}\\\up{2.3}} & \makecell{\textbf{52.0}\\\up{3.7}} & \makecell{\textbf{56.3}\\\UP{5.9}} & \makecell{\textbf{51.2}\\\UP{7.1}} & \makecell{25.0\\\up{4.8}} & \makecell{5.3\\\dn{1.7}} & \makecell{39.6\\\UP{10.3}} & \makecell{\textbf{68.5}\\\UP{5.4}} \\
\bottomrule
\end{tabular}}
\vspace{-8pt}
\end{table*}

Table~\ref{tab:main} shows the success rates of models pretrained with different mixing ratios of Ego2Robot data and real-robot data. 
Table~\ref{tab:breakdown} further compares these models on the extended benchmark, where the test conditions are decomposed along three dimensions: visual appearance, scene, and embodiment. 
From these results, we draw the following conclusions:

\textbf{Co-training improves OOD generalization.}
Ego2R+Robot (1:1) leads in five of seven columns, reaching 53.5\% on RoboTwin Randomized (+2.6 over robot-only) while maintaining 68.1\% on Clean. 
The 1:3 ratio yields marginal gains, whereas 3:1 and 1:1 produce substantial improvements.

\textbf{Visual appearance benefits most.}
Table~\ref{tab:breakdown} shows that 1:1 improves on all three visual factors: background (+4) and lighting (+8) from the scene diversity in egocentric videos, and robot color (+6) from multi-morphology rendering across 15 configurations.

\textbf{Scene layout shows moderate gains.}
Camera offset robustness improves by +6 at 1:1, as egocentric videos naturally exhibit diverse head poses and viewpoints.

\textbf{Embodiment transfer benefits from multi-morphology data.}
ARX improves from 44 to 51 (1:1), and UR5 peaks at 31 (3:1), directly benefiting from exposure to 15 morphologies during pretraining. 
Franka remains below 7\%, reflecting its large kinematic gap from the training embodiment.

\textbf{Task semantics improve consistently.}
Unseen object generalization improves from 29\% to 40\% (+11 at 3:1) due to diverse object interactions in egocentric video. 
Paraphrased instruction robustness reaches 69\% at 1:1, benefiting from broader instruction diversity.

\textbf{EBench confirms gains under higher viewpoint.}
EBench's higher-mounted camera is closer to the egocentric perspective. 
The 3:1 ratio achieves the best EBench score (51.7, +12.1 over robot-only), suggesting co-training benefits are amplified when the viewpoint bias partially matches.

\subsection{Ablation Studies}

\begin{wrapfigure}{r}{0.56\columnwidth}
    \vspace{-12pt}
    \centering
    \includegraphics[width=0.52\columnwidth]{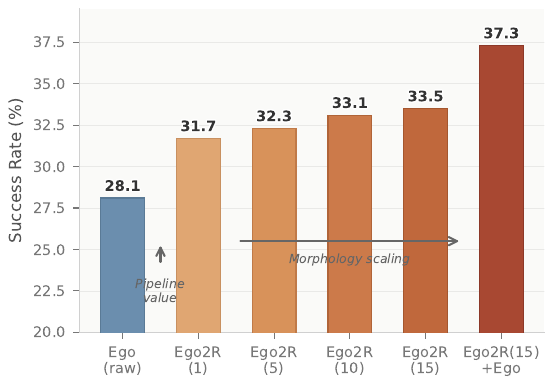}
    \caption{\textbf{Pipeline value and embodiment scaling.} Success rate on RoboTwin Randomized.}
    \label{fig:ablation}
    \vspace{-15pt}
\end{wrapfigure}

Figure~\ref{fig:ablation} isolates the pipeline's contribution using ego-only pretraining (without any robot data).

\textbf{Pipeline alignment is essential.}
Raw ego co-training achieves only 28.1\% on RoboTwin Randomized. 
Processing through our pipeline (Ego2R with a single morphology) improves this to 31.7\%, a +3.6 gain that validates the visual and action alignment stages.

\textbf{More morphologies help.}
Increasing from 1 to 15 morphologies steadily improves performance (31.7$\rightarrow$33.5). Adding raw ego data alongside the 15-morphology Ego2R data yields a further jump to 37.3\%---the raw ego data effectively acts as a 16th ``morphology'' with slightly different visual appearance and action distribution, further enriching the pretraining diversity.

\subsection{Real Robot Experiments}

\begin{figure*}[t]
    \centering
    \includegraphics[width=0.95\textwidth]{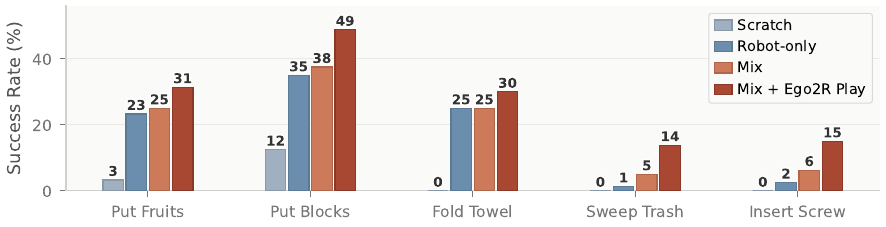}
    \vspace{-8pt}
    \caption{\textbf{Real robot results.} Success rates (\%) on five tasks on the ARX ACone platform.}
    \label{fig:real_results}
    \vspace{-5pt}
\end{figure*}

\begin{figure*}[t]
    \centering
    \includegraphics[width=0.82\textwidth]{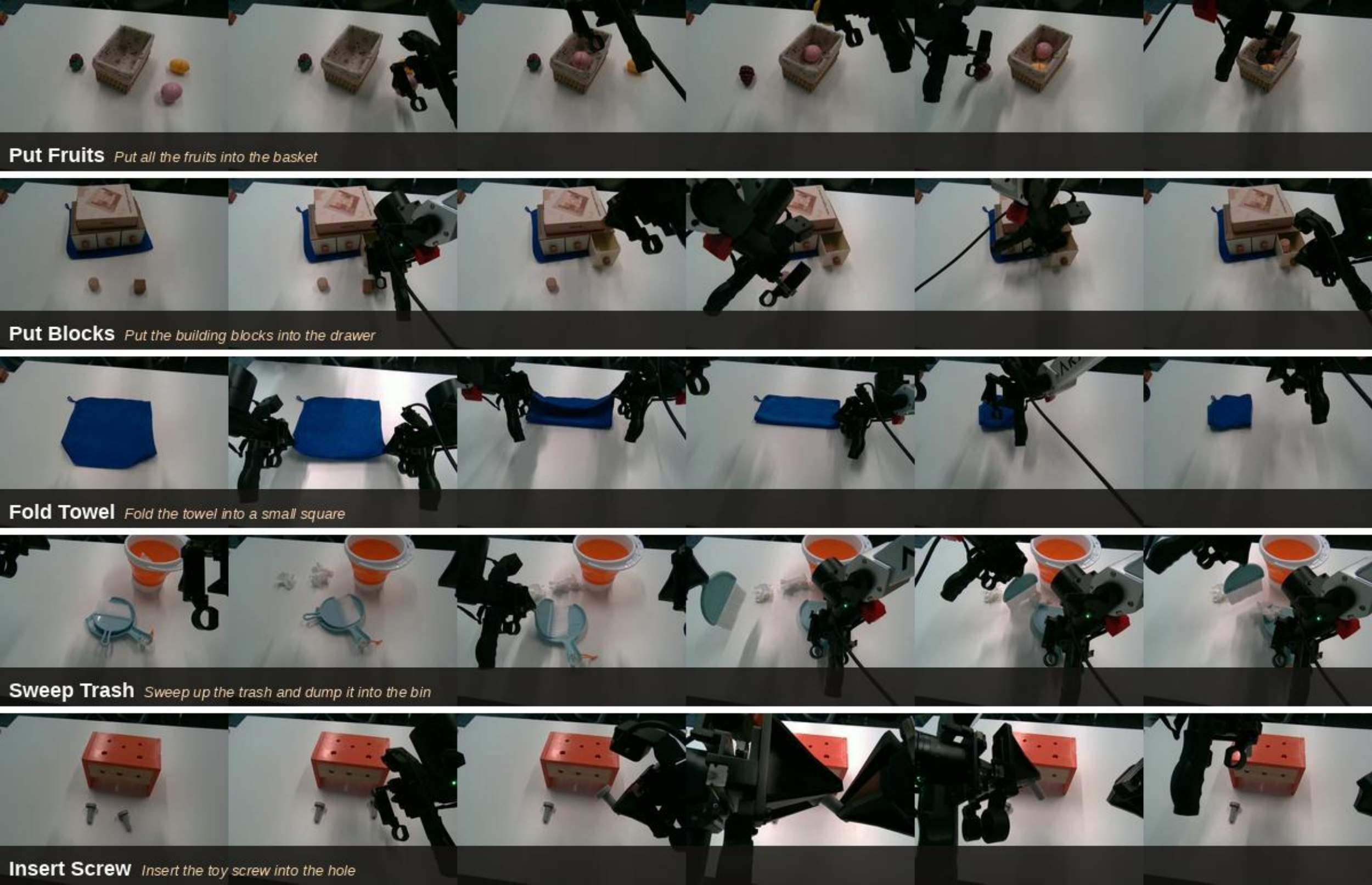}
    \caption{\textbf{Real robot rollouts.} Key frames from five evaluation tasks on the ARX ACone platform.}
    \label{fig:real_tasks}
\end{figure*}

We evaluate on an ARX ACone platform across five long-horizon tasks (Figure~\ref{fig:real_tasks}): putting fruits into a basket, putting blocks into a drawer, folding a towel, sweeping trash into a bin, and inserting a screw. 
For each task, we collect 20 teleoperated demonstrations. Additionally, we record egocentric play videos of hand manipulation (approximately 7 minutes per scene) and process them through our Ego2Robot pipeline to generate ACone-specific synthetic demonstrations. 
We compare three configurations (Figure~\ref{fig:real_results}): \textbf{Robot-only} uses robot-only pretraining and finetunes on teleoperated demonstrations; \textbf{Mix} uses Ego2R+Robot (1:1) pretraining with the same demonstrations; \textbf{Mix + Ego2R Play} finetunes on teleoperated demonstrations mixed 1:1 with the pipeline-converted ego-play (Ego2R) data.

Mix + Ego2R Play achieves the best results across all five tasks, with the largest gains on Put Blocks (+14 over Robot-only) and Insert Screw (+13), demonstrating that casually recorded egocentric videos can be converted into effective training signal via our pipeline. 
Mix pretraining alone already improves over Robot-only, and Ego2R Play data provides further consistent gains.

\section{Limitations}
Our work has several limitations that point to future research directions.
First, retargeting maps hand poses to parallel-jaw grippers, discarding fine-grained finger articulation. Extending to dexterous multi-finger hands could broaden the range of transferable skills.
Second, visual alignment relies on inpainting and depth-aware compositing, which may introduce artifacts under heavy occlusion or complex lighting. Improving rendering fidelity, e.g., with generative models, could further reduce the visual domain gap.
Finally, our evaluation is limited to the task scope of RoboTwin2.0~\cite{chen2025robotwin}. Extending to broader tasks and embodiment configurations would strengthen generality.

\section{Conclusion}

We presented Ego2Robot, a scalable pipeline that converts egocentric hand manipulation videos into robot training data across 15 morphologies. 
The disentangled evaluation we constructed reveals that ego2robot-synthesized data complements robot data, with gains most pronounced under visual, embodiment, and semantic perturbations. 
On real robots, combining ego2robot-synthesized data with a small number of demonstrations enables effective multi-task deployment. 
We hope this work encourages leveraging the vast supply of human manipulation video for scalable robot learning.




\clearpage


\bibliography{corl2026}  

\clearpage

\appendix

\begin{center}
{\LARGE \textbf{Ego2Robot Appendix}}
\end{center}
\vspace{10pt}
\startcontents[appendix]
\printcontents[appendix]{}{1}{\setcounter{tocdepth}{1}}

\section{Ego2Robot Pipeline Details}
\label{app:pipeline}


\subsection{Hand Pose Estimation (Path~B).}

\textit{\textbf{Per-Frame Reconstruction.}}
WiLoR~\cite{potamias2025wilor} performs per-frame hand detection and regresses MANO parameters $(\boldsymbol{\theta}, \boldsymbol{\beta}, \mathbf{t})$, yielding 21 keypoints $\{\mathbf{p}_j\}_{j=0}^{20}$ and a hand mesh per frame.
Detections are filtered by a SAM~3-generated hand mask: frames where more than 80\% of projected keypoints fall outside the mask are discarded.

\textit{\textbf{Cross-Frame Association.}}
WiLoR uses a YOLO-based hand detector that outputs bounding boxes and initial left/right labels. 
To build temporally consistent tracks, the frame with the highest combined left+right detection score serves as the seed. 
From the seed, we propagate bidirectionally: at each frame, each new detection is assigned to the hand whose previous wrist position is nearest in image space ($\ell_2$ distance). 
When multiple detections compete for the same hand, the highest-scoring one is kept. 
After association, a jump filter removes frames where the 3D wrist velocity exceeds $\max(4 \times \text{median velocity}, 0.003\,\text{m/frame})$, indicating misdetections.

\textit{\textbf{Temporal Optimization.}}
DynHaMR~\cite{yu2025dyn} refines the tracks by jointly optimizing MANO pose and shape parameters across the full sequence, minimizing:
\begin{equation}
\small
    \mathcal{L}_\text{dyn} = \mathcal{L}_\text{data} + \lambda_\text{smooth} \mathcal{L}_\text{smooth} + \lambda_\text{bio} \mathcal{L}_\text{bio}
\end{equation}
where $\mathcal{L}_\text{data}$ enforces consistency with per-frame WiLoR estimates, $\mathcal{L}_\text{smooth}$ penalizes acceleration in wrist position and finger joint angles, and $\mathcal{L}_\text{bio}$ enforces biomechanical joint limits. 
The depth of reconstructed hands is constrained to $[0.05, 0.4]$\,m from the camera.

\textit{\textbf{Gap Handling.}}
When hand detections are missing for consecutive frames, we apply gap-specific interpolation. 
Large gaps ($>$10 frames) are filled with the robot's home configuration, with smooth blending transitions at the gap boundaries. 
The blend length is $n = \max(5, \min(90, \lceil 0.6 n_\text{pos} + 0.4 n_\text{rot} \rceil))$ frames, where $n_\text{pos} = \Delta p / 3.25\,\text{mm}$ and $n_\text{rot} = \Delta \theta / 1.08^\circ$ are the number of frames needed at fixed blend speeds to cover the displacement.
Small gaps ($\leq$10 frames) use linear position interpolation + SLERP.

\subsection{Subtask Segmentation (Path~B).}
For long continuous recordings, we pre-split videos at 60-second boundaries. Each clip is sent to a VLM (Qwen3.5~\cite{qwen35blog}) along with the following prompt template:
\begin{quote}
\small
\ttfamily
You are watching a first-person manipulation video (\{duration\}s).\\
Task: "\{task\_desc\}"\\
Segment this video by complete task goals. Each subtask should be an independent, complete objective.\\
Write a SHORT English action instruction (5--12 words) for each subtask.\\
Output ONLY a JSON array: [\{"description":..., "start\_time":..., "end\_time":...\}, ...]
\end{quote}
The VLM returns subtask boundaries and natural language descriptions that serve as training instructions.







\subsection{Action Alignment Details.}

\textit{\textbf{Handedness Sign.}}
The sign $s$ in the grasp-axis definition (Equation~3) is determined by the hand's left/right identity:
$s = +1$ for the right hand and $s = -1$ for the left hand, so that the grasp axis $\mathbf{z}$
points consistently regardless of handedness and both hands map to the same gripper frame.
The wrist-to-fingertip vector $\mathbf{d} = \mathbf{p}_\text{vf} - \mathbf{p}_\text{wrist}$ that forms
the gripper-normal axis $\mathbf{y}$ is taken directly from the MANO keypoints.

\textit{\textbf{Degenerate Orientation.}}
When the thumb tip and virtual fingertip nearly coincide (gripper width $w < 0.01$\,m), the grasp
orientation degenerates; we freeze it to the last valid estimate. The same fallback is applied when
$\mathbf{z}$ and $\mathbf{d}$ are nearly parallel ($\|\mathbf{z} \times \mathbf{d}\| \approx 0$).

\textit{\textbf{Velocity Filtering.}}
Per-frame detection jumps are removed by a velocity filter.
For position velocity $v_t = \|\mathbf{p}_t - \mathbf{p}_{t-1}\|$, the threshold is:
\begin{equation}
\small
    \tau_t = \max\!\bigl(5 \cdot \text{median}(\{v_s\}_{s}), \; 0.9 / \text{fps}\bigr)
\end{equation}
An analogous threshold with floor $10.0/\text{fps}$ rad/frame is used for rotational velocity.
Frames exceeding $\tau_t$ are replaced by interpolation from neighbors. The procedure is iterated for 2 rounds.

\textit{\textbf{Temporal Smoothing.}}
Position and gripper width: Savitzky-Golay filter, window $\min(21, n)$, polynomial order $\min(3, \text{window}{-}1)$.
Orientation: Gaussian-weighted SLERP, $\sigma{=}10$ frames, kernel size 21.
Adjacent quaternions are hemisphere-corrected ($q_t \to -q_t$ when $q_t \cdot q_{t-1} < 0$) before interpolation.


\subsection{Visual Alignment Details.}

\textit{\textbf{Arm Segmentation.}}
SAM~3~\cite{carion2025sam3} uses the text prompt ``person'' with the middle frame as prompt anchor for temporal propagation.
Long videos are processed in 400-frame chunks with 50-frame overlap; masks in overlap regions are merged via bitwise-OR.
Post-processing: (i) gaps $\leq 3$ frames are filled by interpolating neighboring masks; (ii) frames with mask area $< 50\%$ of the local median (window 11) are replaced by the nearest valid mask; (iii) morphological close with $5{\times}5$ elliptical kernel.

\textit{\textbf{Hand Removal.}}
ProPainter~\cite{zhou2023propainter} runs at fp16 with neighbor\_length$\,{=}\,$10, ref\_stride$\,{=}\,$10, subvideo\_length$\,{=}\,$80, mask\_dilation$\,{=}\,$4, RAFT iterations$\,{=}\,$20.

\textit{\textbf{Base Pose Search.}}
The optimization objective is defined in Equation~4. Here we detail the candidate generation and scoring.
Candidates are generated in the camera coordinate frame with offsets scaled by reach $r$:
\begin{itemize}[leftmargin=*, nosep]
    \item Lateral (left-right): $r \times \{0.3, 0.4, 0.5, 0.6, 0.8, 1.0, 1.2\}$ (7 levels, sign-flipped per arm)
    \item Forward-backward: $r \times \{-0.1, 0.0, 0.1, 0.3, 0.5, 0.7, 0.9\}$ (7 levels)
    \item Vertical: $r \times \{0.4, 0.2, 0.0, -0.2, -0.4\}$ (5 levels)
    \item Orientation: pitch $\{30^\circ, 45^\circ, 60^\circ\}$ $\times$ yaw $\{-45^\circ, -20^\circ, 0^\circ, 20^\circ, 45^\circ\}$ $\times$ roll $\{-15^\circ, 0^\circ, 15^\circ\}$
\end{itemize}
Candidates with camera distance $< 0.20$\,m, or with trajectory points beyond $0.9r$ from base or closer than 0.08\,m to base, are discarded.
Surviving candidates are scored by:
\begin{equation}
\small
    S = \text{FR}(\mathbf{T}_\text{base}) - 5.0 \cdot |\bar{\rho} - 0.65|
\end{equation}
where $\text{FR}$ is the IK feasibility rate over up to 20 keyframes (using the mink IK solver, quadprog backend, 100 iterations, $10^{-5}$ threshold). 
The second term is a reach penalty: $\bar{\rho} = \frac{1}{|\mathcal{K}|}\sum_{k} \|\mathbf{T}_\text{base}^{-1}\mathbf{p}_k^\text{ee}\| / r$ is the average end-effector distance from the base normalized by the kinematic reach $r$, and the penalty is minimized when $\bar{\rho} = 0.65$, encouraging the robot to operate at 65\% of its maximum reach to retain kinematic margin for manipulation.
We independently screen the top-5 candidates per arm, then jointly verify all 25 left--right combinations to select the final base placement.

\textit{\textbf{Depth-Aware Compositing.}}
The main text (Equation~5) presents a unified depth-ordering rule. 
In practice, we separate the robot into arm body and gripper regions using MuJoCo's per-geom segmentation masks. 
The arm body mask ($\texttt{robot\_mask}{=}1 \wedge \texttt{gripper\_mask}{=}0$) is always composited onto the inpainted scene, since the arm is positioned between the camera and the workspace and is never occluded by scene objects. 
The gripper mask ($\texttt{gripper\_mask}{=}1$) undergoes per-pixel depth comparison: a gripper pixel at $(u,v)$ is hidden when the scene depth $D_\text{scene}(u,v) < D_\text{sim}(u,v)$ \emph{and} the pixel is not within the dilated hand mask region, where $D_\text{scene}$ is estimated by Depth Anything V3 and $D_\text{sim}$ is rendered by MuJoCo. 
The hand mask is dilated with a $5{\times}5$ kernel (1 iteration) before compositing to prevent revealing inpainted boundaries along the original arm contour. The final overlay is:
\begin{equation}
\small
    I_\text{out}(u,v) = \begin{cases}
        I_\text{sim}(u,v) & \text{if } (u,v) \in M_\text{arm} \cup M_\text{gripper\_vis} \\
        I_\text{inpaint}(u,v) & \text{otherwise}
    \end{cases}
\end{equation}
where $M_\text{gripper\_vis} = M_\text{gripper} \setminus \{(u,v) : D_\text{scene} < D_\text{sim} \wedge (u,v) \notin M_\text{hand}\}$.


\subsection{Quality Curation Details.}

\textit{\textbf{L1 (Pipeline-internal).}}
A frame is valid when: (i) hand detected, (ii) IK tracking error $< 0.05$\,m, (iii) rendered robot visible ($>$0 pixels), (iv) no self-collision (MuJoCo contact check).
Additional invalidation: bimanual cross-arm contacts $> 1$ or robot mask $> 70\%$ of image area.

\textit{\textbf{Stability Erosion.}}
Short valid runs ($< \lfloor 0.3 \times \text{fps} \rfloor$ frames) sandwiched between invalid regions are eroded to invalid.

\textit{\textbf{L2 (Statistical).}}
Two post-hoc statistical filters are applied:
(i) \emph{Q1/Q99 filter}: for each action dimension, per-dataset Q1 and Q99 statistics are computed (excluding previously invalidated frames). Frames falling outside $[\text{Q1} - 3(\text{Q99}-\text{Q1}),\; \text{Q99} + 3(\text{Q99}-\text{Q1})]$ are flagged.
(ii) \emph{Sudden-change filter}: for each step, residual, acceleration, and jerk are computed on state/action trajectories. Steps exceeding per-dimension thresholds (computed from the dataset's distribution) are flagged.
After both filters, episodes whose total invalid frame ratio exceeds 60\% are discarded entirely. 

\textit{\textbf{L3 (VLM Consistency).}}
A VLM (Qwen3.5) audits each synthesized video for semantic consistency. The video is sampled at 4\,fps and sent along with the subtask description. 
The prompt is:
\begin{quote}
\small\ttfamily
You are evaluating whether a manipulation video matches its text description.\\[2pt]
This is a ROBOT manipulation dataset. "Hand" refers to the robot's gripper/end-effector.\\
Many tasks use FAKE or SIMULATED objects as stand-ins for real objects. This is EXPECTED.\\[2pt]
Task Description: \{description\}\\[2pt]
Determine if the robot's actions match the described task.\\
Flag MAJOR MISMATCHES: wrong action type, wrong object category, wrong target location, or failed execution.\\
Be tolerant of: fake/toy objects, minor appearance variations, small spatial deviations, different grasping approaches.\\[2pt]
Respond in JSON: \{"is\_consistent": true/false, "confidence": 0.0-1.0, "reasoning": "..."\}
\end{quote}
Episodes judged inconsistent are discarded.




\subsection{Color Randomization.}
Each rendered robot undergoes deterministic color randomization in HSV space: $H \sim U(0, 1)$, $S \sim U(0.3, 1)$, $V \sim U(0.4, 1)$, applied uniformly across all links.


\subsection{Supported Robot Morphologies.}
\begin{figure*}[t]
\centering
\includegraphics[width=\textwidth]{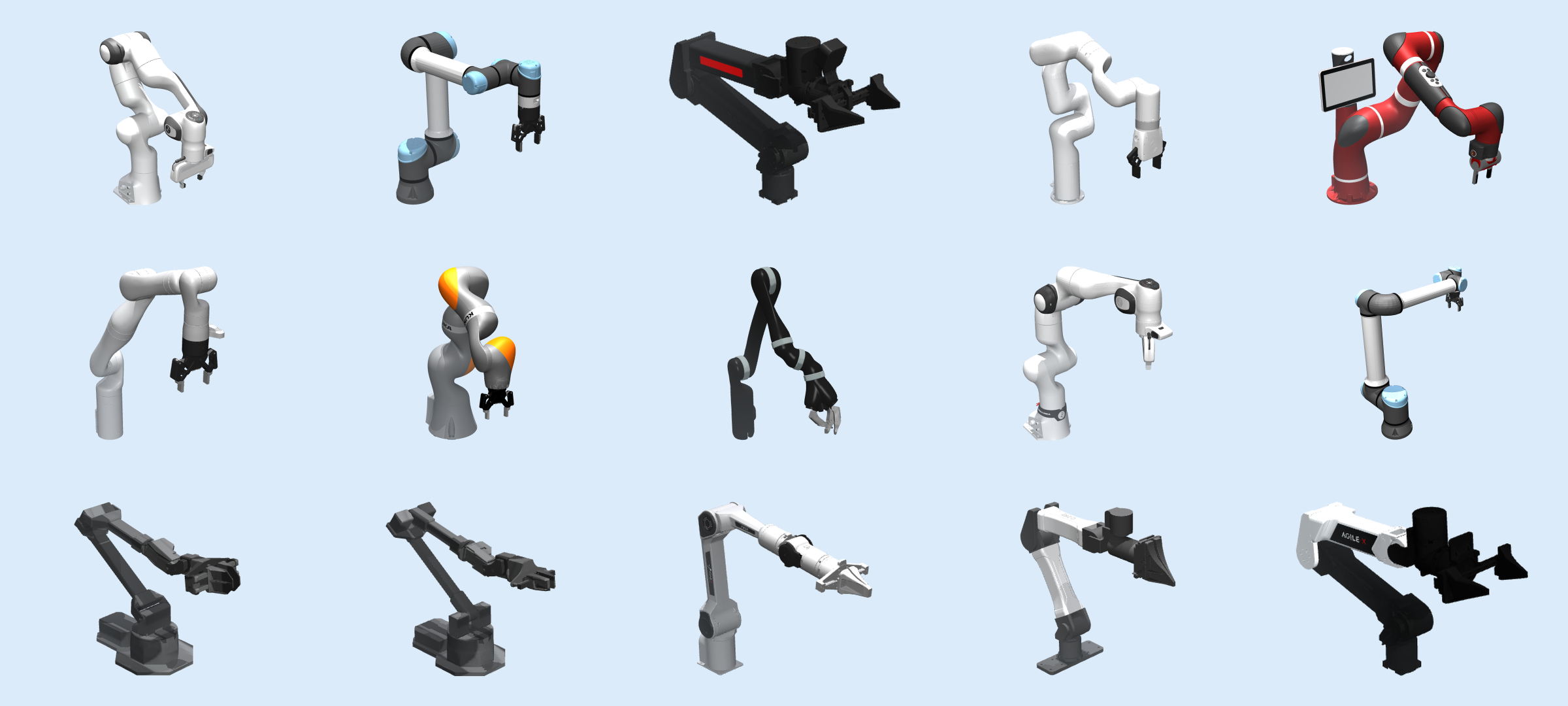}
\caption{\textbf{15 supported robot morphologies.} 3D models of all 15 robots.}
\label{fig:robot_gallery}
\vspace{-5pt}
\end{figure*}

Table~\ref{tab:robots} lists the 15 morphologies. Figure~\ref{fig:robot_gallery} shows the 3D models of all 15 robots.

\begin{table}[t]
\centering
\small
\caption{\textbf{Supported robot morphologies.}}
\label{tab:robots}
\begin{tabular}{lccc}
\toprule
Robot & DOF & Gripper (mm) & Reach (m) \\
\midrule
Panda       & 7 & 0--80  & 1.272 \\
Kinova Gen3 & 7 & 0--85  & 1.337 \\
IIWA        & 7 & 0--85  & 1.411 \\
Sawyer      & 7 & 14--79 & 1.420 \\
FR3         & 7 & 0--80  & 1.272 \\
xArm7       & 7 & 0--85  & 1.290 \\
UR5e        & 6 & 0--85  & 1.236 \\
UR10e       & 6 & 0--85  & 1.627 \\
Jaco        & 6 & 0--125 & 1.200 \\
ViperX      & 6 & 15--87 & 0.911 \\
WidowX      & 6 & 11--55 & 0.787 \\
ARX-L5      & 6 & 0--88  & 0.855 \\
Piper       & 6 & 0--70  & 0.883 \\
YAM         & 6 & 4--75  & 0.866 \\
Aloha-Agilex& 6 & 7--102 & 0.853 \\
\bottomrule
\end{tabular}
\end{table}

\section{Evaluation Framework Details}
\label{app:benchmark}

\subsection{Perturbation Parameters.}

\textit{\textbf{Randomized (standard OOD).}} The original RoboTwin 2.0 setting simultaneously applies background texture replacement, lighting randomization, table height offset, and clutter. 
We decouple these into independent axes below, and additionally introduce Camera Offset, Robot Color, Paraphrased Instructions, and Unseen Objects as new evaluation dimensions.

\textit{Visual Appearance:}
\textit{\textbf{Background.}} Table and floor textures replaced with randomly sampled textures from a library.
\textit{\textbf{Lighting.}} Light positions and intensities randomized.
\textit{\textbf{Robot Color.}} Uniform hue shift from $[0^\circ, 360^\circ)$ applied to all robot links.

\textit{Scene Layout:}
\textit{\textbf{Table Height.}} Vertical offset $\pm$4\,cm.
\textit{\textbf{Clutter.}} 3--5 distractor objects with random poses.
\textit{\textbf{Camera Offset.}} Head camera displaced up to 5\,cm per axis.

\textit{Task Semantics:}
\textit{\textbf{Paraphrased Instructions.}} 505 colloquially rephrased commands (human + LLM generated).
\textit{\textbf{Unseen Objects.}} 50 new tasks with object instances not present in the finetuning demonstrations (same category, different appearance/geometry), spanning seven manipulation families:
\begin{itemize}[leftmargin=*, nosep]
    \item \textbf{Move-to-pad/pot} (10): \emph{move bottle pad, move cup pad, move bowl pad, move cup pot, move bottle pot, move mug pot, move apple pot, move seal pot, move pillbottle pot, move bowl pot}
    \item \textbf{Place-in-basket} (11): \emph{place soap basket, place mug basket, place perfume basket, place teabox basket, place toycar basket, place apple basket, place bottle basket, place cup basket, place coffeebox basket, place saucecan basket, place hamburg basket}
    \item \textbf{Place-on-stand} (4): \emph{place cup stand, place mug stand, place bottle stand, place apple stand}
    \item \textbf{Place-on-scale} (3): \emph{place bottle scale, place cup scale, place mug scale}
    \item \textbf{Put-in-cabinet} (3): \emph{put apple cabinet, put seal cabinet, put bowl cabinet}
    \item \textbf{Bimanual place-A-to-B} (12): \emph{place cup left bottle, place mug right apple, place bottle left cup, place seal right mug, place mug left cup, place apple left mug, place bottle right apple, place seal left cup, place mug left bottle, place apple right cup, place cup right seal, place bottle left apple}
    \item \textbf{Move-away} (7): \emph{move cup away, move bottle away, move apple away, move mug away, move seal away, move bowl away, move pillbottle away}
\end{itemize}

\textit{Embodiment:}
\textit{\textbf{Cross-Embodiment Transfer.}} The default embodiment is Aloha-Agilex, loaded as a single dual-arm URDF. 
For zero-shot cross-embodiment evaluation, we replace it with ARX-X5, UR5-WSG, or Franka Panda. 
Each alternative embodiment is loaded as two independent single-arm URDFs positioned symmetrically about the workspace center, with inter-arm base distances of 0.6\,m (ARX-X5), 0.59\,m (UR5-WSG), and 0.65\,m (Franka) to accommodate different arm lengths. 
Initial EEF poses are aligned via IK, and all three camera extrinsics (head camera + two wrist cameras) are aligned across embodiments to ensure consistent starting conditions in the camera-frame action space.

\subsection{Task List.}
The 50 tasks: \emph{adjust bottle, beat block hammer, blocks ranking rgb, blocks ranking size, click alarmclock, click bell, dump bin bigbin, grab roller, handover block, handover mic, hanging mug, lift pot, move can pot, move pillbottle pad, move playingcard away, move stapler pad, open laptop, open microwave, pick diverse bottles, pick dual bottles, place a2b left, place a2b right, place bread basket, place bread skillet, place burger fries, place can basket, place cans plasticbox, place container plate, place dual shoes, place empty cup, place fan, place mouse pad, place object basket, place object scale, place object stand, place phone stand, place shoe, press stapler, put bottles dustbin, put object cabinet, rotate qrcode, scan object, shake bottle, shake bottle horizontally, stack blocks three, stack blocks two, stack bowls three, stack bowls two, stamp seal, turn switch.}

Step limits per task range from 400 (e.g., click bell, adjust bottle) to 1700 (put bottles dustbin), with most tasks at 400. Complex multi-step tasks use higher limits: open microwave 1500, blocks ranking 1200, stack blocks three 1200, stack bowls 900--1200, hanging mug 900, stack blocks two / handover block 800, place cans plasticbox 800. 
New/unseen tasks default to 1000 steps.

\subsection{Protocol.}
50 episodes/task, binary success.
Aggregates: Visual = mean(BG, Light, Robot Color); Scene = mean(Height, Clutter, Camera); Embodiment = mean(ARX, UR5, Franka); Task = mean(Obj, Lang).

\subsection{EBench Table Top.}
EBench~\cite{ebench2026} provides 7 table-top tasks in Isaac Sim: \emph{collect coffee beans, flip cup \& collect cookies, frame against pen holder, install gear, peg in hole, put glass in glass box, tighten nut}. Camera is higher than RoboTwin, closer to the egocentric perspective.

\section{Model Architecture Details}
\label{app:model}

The Qwen3.5-4B vision-language backbone processes multi-view RGB images (2--3 views per timestep) and a structured text prompt, producing a sequence of hidden states $\mathbf{H} \in \mathbb{R}^{L \times d}$ that jointly encode visual and linguistic information. 
Camera intrinsics $\mathbf{K} \in \mathbb{R}^{3 \times 3}$ and extrinsics $\mathbf{T}_{wc} \in SE(3)$ are injected into the visual features via mRoPE positional encodings, enabling the model to reason about the 3D spatial relationship between the camera and the scene. 
The DiT action head conditions on $\mathbf{H}$ through cross-attention: a context generator compresses $\mathbf{H}$ into conditioning features $\mathbf{C}$ using 8 learnable query tokens, and the DiT layers attend to $\mathbf{C}$ when predicting 32-step action chunks via flow matching (8 training / 4 inference steps).

\subsection{Structured Prompt.}
The text input follows a two-field format:
\begin{verbatim}
embodiment: {type}_{model}
instruction: {task description}
\end{verbatim}
\texttt{embodiment} encodes data source and robot model: \texttt{robot\_aloha} (real robot), \texttt{h2r\_arx} (Ego2R with ARX), \texttt{human\_ego} (raw ego). Dropped to ``None'' with 15\% probability during training. \texttt{instruction} is never dropped.

\subsection{Camera-Frame Relative EEF.}
Each action step is 7-dim: 3D position delta $\Delta\mathbf{p}$, 3D rotation delta $\Delta\boldsymbol{\omega}$ (rotation vector), 1D gripper.
Given the EEF-to-camera transformation $\mathbf{T}_{ce} = \mathbf{T}_{wc}^{-1} \mathbf{T}_{we}$ with rotation component $\mathbf{R}_{ce}$, base-frame deltas are transformed:
\begin{equation}
\small
    \Delta\mathbf{p}_{cc} = \mathbf{R}_{ce}\, \Delta\mathbf{p}_{ee}, \quad
    \Delta\mathbf{R}_{cc} = \mathbf{R}_{ce}\, \Delta\mathbf{R}_{ee}\, \mathbf{R}_{ce}^\top
\end{equation}
This naturally unifies data from different camera setups and robot morphologies without explicit extrinsic calibration, as described in Section~\ref{sec:pipeline}. 

\subsection{Flow Matching.}
Given clean actions $\mathbf{a}_0$ and noise $\boldsymbol{\epsilon} \sim \mathcal{N}(\mathbf{0}, \mathbf{I})$, the noised sample at $t \in [0,1]$:
\begin{equation}
\small
    \mathbf{a}_t = (1-t)\,\mathbf{a}_0 + t\,\boldsymbol{\epsilon}
\end{equation}
Training objective:
\begin{equation}
\small
    \mathcal{L} = \mathbb{E}_{t, \mathbf{a}_0, \boldsymbol{\epsilon}} \left\| \mathbf{v}_\theta(\mathbf{a}_t, t, \mathbf{c}) - (\boldsymbol{\epsilon} - \mathbf{a}_0) \right\|^2
\end{equation}
where $\mathbf{c}$ is the visual-language conditioning. Inference uses 4 Euler steps ($\Delta t {=} 1/4$).

\section{Training Details}
\label{app:training}

\subsection{Pretraining.}
8$\times$A100 GPUs, batch 12/GPU, backbone lr $10^{-5}{\to}10^{-6}$ cosine (5K warmup), action head lr $10\times$ ($10^{-4}{\to}10^{-5}$), AdamW ($\beta_1{=}0.9, \beta_2{=}0.95$), bf16, ZeRO-1, 200K steps ($\sim$19.2M frames), 8 diffusion steps/forward. 
No image augmentation.

\subsection{Pretraining Data.}
\begin{figure}[t]
\centering
\includegraphics[width=0.85\columnwidth]{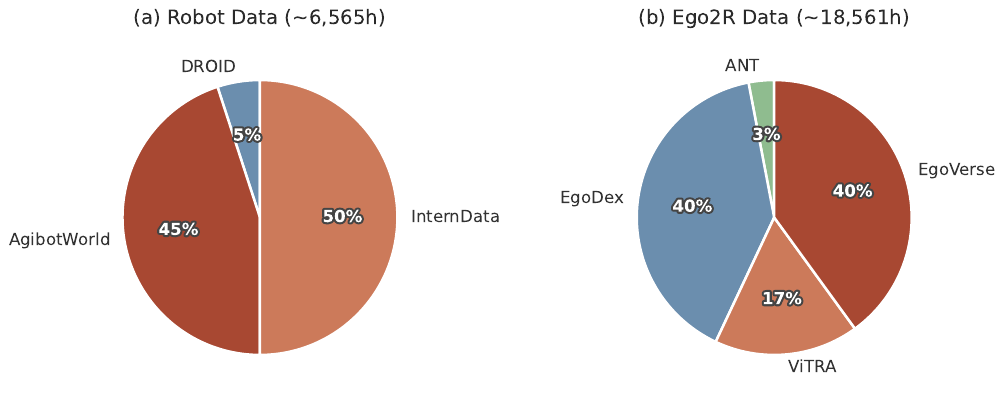}
\caption{\textbf{Pretraining data composition.} 
Slices show per-source \emph{sampling weights} (training mix); panel titles give the total data volume. 
(a) Robot ($\sim$6{,}565\,h). 
(b) Ego2R ($\sim$18{,}561\,h).}
\label{fig:data_pie}
\end{figure}

Robot data comprises three sources:
DROID~\cite{khazatsky2024droid} ($\sim$511\,h) provides real-world teleoperated demonstrations across diverse objects and scenes with a Franka arm.
AgibotWorld~\cite{bu2025agibot} ($\sim$2,404\,h) is a large-scale real teleoperation dataset covering 180 manipulation tasks on humanoid platforms.
InternData~\cite{tian2025interndata} ($\sim$3,650\,h) provides simulation-generated demonstrations across Franka, humanoid, and Aloha embodiments.
Together these total $\sim$6,565\,h.
Ego2R data is derived from four egocentric sources (ANT 7\,h, EgoDex 732\,h, ViTRA 249\,h, EgoVerse 954\,h, totaling $\sim$1,940\,h before processing) rendered across 15 morphologies, yielding 18,561\,h after pipeline processing and quality curation.
The mixing ratios (1:3, 3:1, 1:1 Ego2R-to-Robot) are implemented via per-source sampling weights; all configurations process the same total number of frames. Figure~\ref{fig:data_pie} shows the per-source sampling weights used within each group.

\subsection{Finetuning.}
All finetuning runs load pretrained weights. 8$\times$A100 GPUs, batch 12/GPU, backbone lr $10^{-5}{\to}10^{-6}$ cosine (5K warmup), action head lr $10\times$ ($10^{-4}{\to}10^{-5}$), AdamW ($\beta_1{=}0.9, \beta_2{=}0.95$), bf16, ZeRO-2, ColorJitter, 8 diffusion steps/forward.

\textit{RoboTwin Clean.} 50 tasks $\times$ 50 demos (2,500 episodes), 50K steps, chunk 20, replan every 16.

\textit{EBench Table Top.} 7 table-top tasks $\times$ 400 demos (2,800 episodes), 50K steps, chunk 32.

\textit{Real robot.} See Section~\ref{app:real} for data details. 1:1 teleop-to-Ego2R mixing ratio, 50K steps, chunk 32.

\subsection{Ablation Configurations.}
The ablation (Figure~\ref{fig:ablation}) uses only ego-sourced data (raw ego or Ego2R) for pretraining, excluding real robot data, to isolate the pipeline's contribution:
\begin{itemize}[leftmargin=*, nosep]
    \item \textbf{Raw ego} ($\sim$1,940\,h): egocentric video after pipeline quality filtering but without robot rendering.
    \item \textbf{Ego2R, 1 morphology} ($\sim$1,237\,h): pipeline-processed, ARX-L5 only.
    \item \textbf{Ego2R, 5 morphologies} ($\sim$6,187\,h): Aloha-Agilex, ARX-L5, FR3, ViperX, xArm7.
    \item \textbf{Ego2R, 10 morphologies} ($\sim$12,374\,h): Aloha-Agilex, ARX-L5, FR3, IIWA, Jaco, Piper, UR5e, ViperX, YAM, xArm7.
    \item \textbf{Ego2R, 15 morphologies} ($\sim$18,561\,h): all 15.
    \item \textbf{Ego2R (15) + Raw ego} ($\sim$20,501\,h): 15-morphology Ego2R plus raw egocentric data.
\end{itemize}
All ablation models are pretrained on the specified ego-sourced data only (no real robot data), then finetuned on RoboTwin Clean.

\begin{figure*}[t]
\centering
\includegraphics[width=\textwidth]{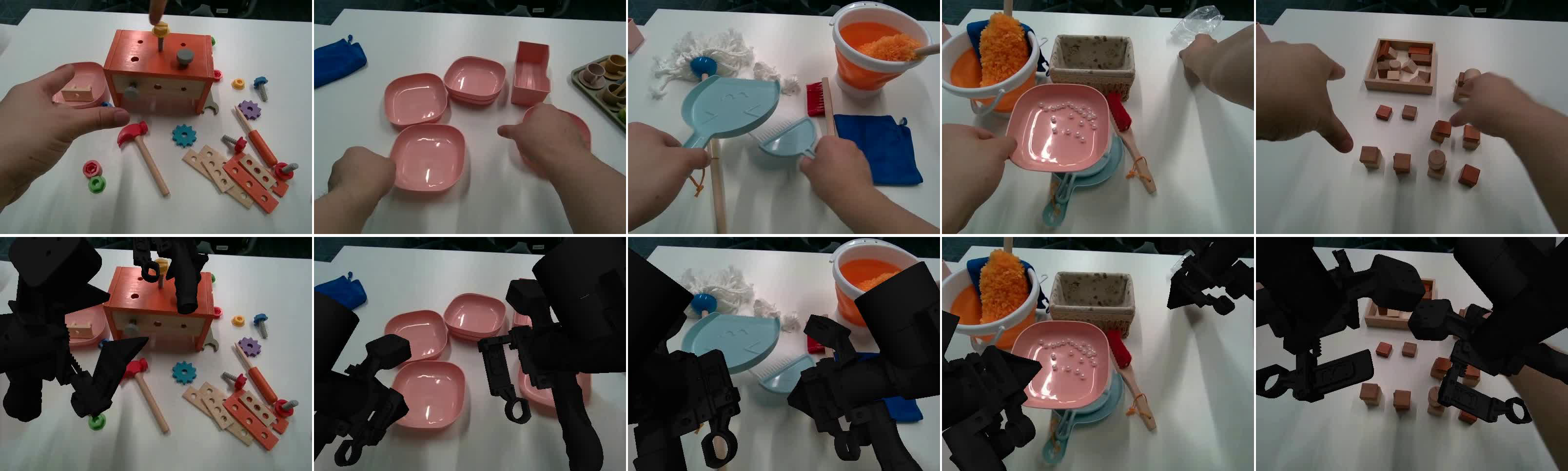}
\caption{\textbf{Ego play $\to$ Ego2R synthesis on real-world scenes.} Top: original egocentric human manipulation. Bottom: Ego2R pipeline output with ACone robot overlay. 
}
\label{fig:ego2r_realplay}
\end{figure*}

\section{Real Robot Experiment Details}
\label{app:real}

\subsection{Platform and Tasks.}
ARX ACone dual-arm (6-DOF/arm, parallel gripper, head + two wrist RGB cameras, 15\,Hz control, Replan every 32 steps).
\begin{itemize}[leftmargin=*, nosep]
    \item \textbf{Put Fruits} (3 steps): place 3 fruits into a basket.
    \item \textbf{Put Blocks} (4 steps): open drawer, place 2 blocks, close drawer.
    \item \textbf{Fold Towel} (2 steps): two sequential folds.
    \item \textbf{Sweep Trash} (4 steps): pick broom, sweep, dump, return broom.
    \item \textbf{Insert Screw} (4 steps): bimanual handover and insertion of 2 screws.
\end{itemize}

\subsection{Data.}
20 teleop demos per task (100 total) + $\sim$35\,min ego play video across 5 scenes. Ego play processed via Path~B (WiLoR, DynHaMR, Qwen3.5, Ego2R) to generate 675 ACone synthetic episodes. 
Teleop and Ego2R data are mixed at a 1:1 ratio during finetuning.

\subsection{Ego Play to Ego2R Visualization.}
Figure~\ref{fig:ego2r_realplay} shows representative frames from the ego play data and corresponding Ego2R synthesis. 
The top row shows the original egocentric human manipulation; the bottom row shows the pipeline output with the robot (ACone) overlaid onto the inpainted scene.


\subsection{Scoring.}
Sub-step partial scoring (Table~\ref{tab:scoring}), 100 points per task, 20 trials.

\begin{table}[t]
\centering
\small
\caption{\textbf{Real robot scoring.}}
\label{tab:scoring}
\begin{tabular}{lcccc}
\toprule
Task & Step 1 & Step 2 & Step 3 & Step 4 \\
\midrule
Put Fruits   & 33.3 & 33.3 & 33.3 & --- \\
Put Blocks   & 25   & 25   & 25   & 25 \\
Fold Towel   & 50   & 50   & ---  & --- \\
Sweep Trash  & 25   & 25   & 25   & 25 \\
Insert Screw & 25   & 25   & 25   & 25 \\
\bottomrule
\end{tabular}
\end{table}

\section{Additional Experimental Results}
\label{app:results}

\subsection{Comparison with Pi0.5.}
\begin{figure}[t]
\centering
\includegraphics[width=\columnwidth]{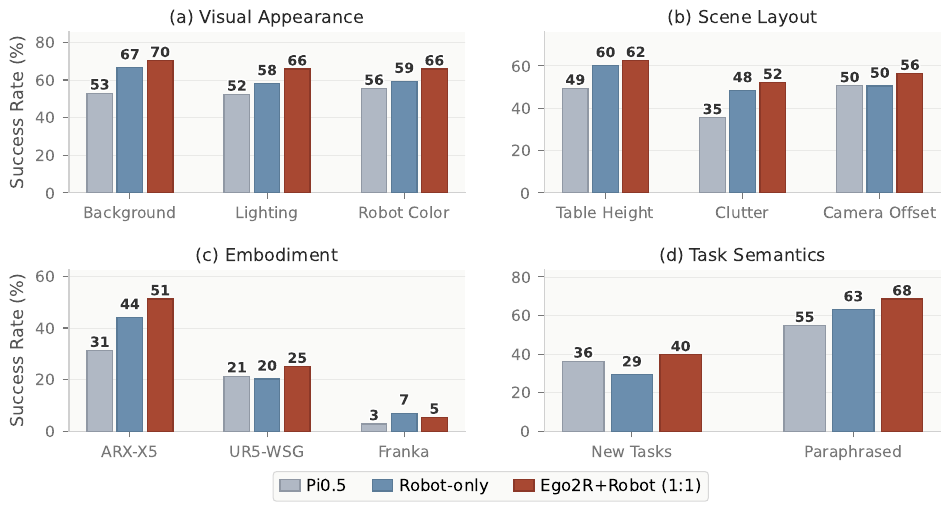}
\caption{\textbf{Comparison with Pi0.5} across RoboTwin settings.}
\label{fig:pi05}
\end{figure}

Pi0.5~\cite{intelligence2025pi05} is evaluated under the same RoboTwin EEF settings (OpenPI framework, camera-frame relative EEF). Pi0.5 uses a different backbone and pretraining data. Figure~\ref{fig:pi05} reports results. Ego2R+Robot (1:1) outperforms the Robot-only baseline across nearly all settings.

\subsection{Per-Task Results.}
Table~\ref{tab:pertask} reports per-task success rates on all 50 RoboTwin tasks under Clean and Randomized settings for five models: Pi0.5 and four of our models.

\begin{table*}[t]
\centering
\small
\setlength{\tabcolsep}{2.5pt}
\caption{\textbf{Per-task results on RoboTwin including Pi0.5} (success rate \%).}
\label{tab:pertask}
\begin{tabular}{l|cc|cc|cc|cc|cc}
\toprule
\multirow{2}{*}{Task} & \multicolumn{2}{c|}{Pi0.5} & \multicolumn{2}{c|}{Robot-only} & \multicolumn{2}{c|}{Ego2R (1:3)} & \multicolumn{2}{c|}{Ego2R (3:1)} & \multicolumn{2}{c}{Ego2R (1:1)} \\
 & Clean & Rand & Clean & Rand & Clean & Rand & Clean & Rand & Clean & Rand \\
\midrule
adjust\_bottle             & 62.0 & 18.0 & 94.0 & 77.0 & 92.0 & 79.0 & \textbf{100.0} & 72.0 & 89.0 & \textbf{91.0} \\
beat\_block\_hammer        & 48.0 & 16.0 & 63.0 & 27.0 & 64.0 & 56.0 & 57.0 & 18.0 & \textbf{65.0} & \textbf{71.0} \\
blocks\_ranking\_rgb       & 74.0 & 16.0 & 56.0 & 60.0 & 64.0 & \textbf{63.0} & 49.0 & 43.0 & \textbf{78.0} & 60.0 \\
blocks\_ranking\_size      & 32.0 & 0.0 & 30.0 & \textbf{22.0} & 46.0 & 21.0 & 43.0 & 7.0 & \textbf{52.0} & 16.0 \\
click\_alarmclock          & 44.0 & 56.0 & 96.0 & 93.0 & \textbf{100.0} & 87.0 & \textbf{100.0} & 85.0 & \textbf{100.0} & \textbf{100.0} \\
click\_bell                & 32.0 & 40.0 & \textbf{100.0} & 95.0 & \textbf{100.0} & 94.0 & \textbf{100.0} & 93.0 & \textbf{100.0} & \textbf{100.0} \\
dump\_bin\_bigbin           & 72.0 & 70.0 & 54.0 & 73.0 & 72.0 & 69.0 & 81.0 & \textbf{76.0} & \textbf{86.0} & 72.0 \\
grab\_roller               & 94.0 & 46.0 & \textbf{98.0} & \textbf{71.0} & 68.0 & 48.0 & 91.0 & 64.0 & 89.0 & 63.0 \\
handover\_block            & 16.0 & 0.0 & 5.0 & 2.0 & 18.0 & 4.0 & \textbf{39.0} & \textbf{9.0} & 36.0 & \textbf{9.0} \\
handover\_mic              & 26.0 & 4.0 & 69.0 & 13.0 & 86.0 & \textbf{34.0} & 83.0 & 17.0 & \textbf{97.0} & 23.0 \\
hanging\_mug               & 10.0 & 4.0 & \textbf{14.0} & \textbf{16.0} & \textbf{14.0} & 9.0 & 12.0 & 9.0 & 10.0 & 10.0 \\
lift\_pot                  & 8.0 & 2.0 & 93.0 & 28.0 & 84.0 & 14.0 & \textbf{95.0} & \textbf{44.0} & 93.0 & \textbf{44.0} \\
move\_can\_pot             & 28.0 & 0.0 & \textbf{47.0} & 50.0 & 30.0 & 40.0 & 42.0 & \textbf{85.0} & \textbf{47.0} & 65.0 \\
move\_pillbottle\_pad      & 60.0 & 44.0 & 63.0 & 60.0 & 68.0 & 69.0 & 41.0 & 67.0 & \textbf{74.0} & \textbf{82.0} \\
move\_playingcard\_away    & 90.0 & 52.0 & 74.0 & 58.0 & 88.0 & \textbf{92.0} & 88.0 & 42.0 & \textbf{92.0} & 63.0 \\
move\_stapler\_pad         & 22.0 & 6.0 & 23.0 & 19.0 & 34.0 & 21.0 & 30.0 & 13.0 & \textbf{43.0} & \textbf{31.0} \\
open\_laptop               & 68.0 & 20.0 & \textbf{77.0} & \textbf{61.0} & 74.0 & 56.0 & 64.0 & 46.0 & 75.0 & 58.0 \\
open\_microwave            & 26.0 & 8.0 & \textbf{77.0} & \textbf{59.0} & 42.0 & 29.0 & 37.0 & 32.0 & 50.0 & 25.0 \\
pick\_diverse\_bottles     & 56.0 & 18.0 & 60.0 & 37.0 & 58.0 & \textbf{53.0} & \textbf{71.0} & 50.0 & 70.0 & 50.0 \\
pick\_dual\_bottles        & 82.0 & 14.0 & 91.0 & 59.0 & 80.0 & \textbf{77.0} & 83.0 & 55.0 & \textbf{97.0} & 54.0 \\
place\_a2b\_left           & 60.0 & 10.0 & 40.0 & \textbf{55.0} & 58.0 & 42.0 & 54.0 & 48.0 & \textbf{73.0} & 44.0 \\
place\_a2b\_right          & 58.0 & 14.0 & 42.0 & 49.0 & 52.0 & 41.0 & 53.0 & 51.0 & \textbf{64.0} & \textbf{53.0} \\
place\_bread\_basket       & 68.0 & 44.0 & 75.0 & 55.0 & 72.0 & \textbf{64.0} & 76.0 & 59.0 & \textbf{79.0} & 62.0 \\
place\_bread\_skillet      & 86.0 & 46.0 & 76.0 & 41.0 & 76.0 & 50.0 & 80.0 & \textbf{61.0} & \textbf{87.0} & 53.0 \\
place\_burger\_fries       & 90.0 & 78.0 & 96.0 & 83.0 & \textbf{98.0} & 81.0 & 97.0 & \textbf{88.0} & 96.0 & \textbf{88.0} \\
place\_can\_basket         & 40.0 & 0.0 & \textbf{49.0} & \textbf{25.0} & 38.0 & 9.0 & 34.0 & 15.0 & 34.0 & 16.0 \\
place\_cans\_plasticbox    & 94.0 & \textbf{74.0} & 90.0 & 68.0 & 54.0 & 68.0 & \textbf{97.0} & 59.0 & 70.0 & 59.0 \\
place\_container\_plate    & \textbf{96.0} & 58.0 & 86.0 & 76.0 & 92.0 & 79.0 & 93.0 & 73.0 & 93.0 & \textbf{81.0} \\
place\_dual\_shoes         & \textbf{54.0} & 12.0 & 50.0 & \textbf{30.0} & 34.0 & 27.0 & 35.0 & 17.0 & 35.0 & 19.0 \\
place\_empty\_cup          & 74.0 & 56.0 & 73.0 & 74.0 & 86.0 & \textbf{83.0} & \textbf{88.0} & 78.0 & \textbf{88.0} & 66.0 \\
place\_fan                 & 36.0 & 30.0 & \textbf{48.0} & \textbf{54.0} & \textbf{48.0} & 24.0 & 41.0 & 50.0 & 39.0 & 46.0 \\
place\_mouse\_pad          & 18.0 & 0.0 & \textbf{41.0} & 34.0 & 22.0 & \textbf{36.0} & 30.0 & 25.0 & 29.0 & 33.0 \\
place\_object\_basket      & 24.0 & 4.0 & 65.0 & 50.0 & 76.0 & 28.0 & \textbf{77.0} & \textbf{51.0} & 69.0 & 48.0 \\
place\_object\_scale       & \textbf{58.0} & 44.0 & 38.0 & 46.0 & 40.0 & 33.0 & 52.0 & 45.0 & 40.0 & \textbf{47.0} \\
place\_object\_stand       & \textbf{94.0} & 40.0 & 70.0 & 75.0 & 80.0 & 77.0 & 87.0 & 71.0 & 82.0 & \textbf{82.0} \\
place\_phone\_stand        & 40.0 & 4.0 & 65.0 & 49.0 & 48.0 & \textbf{63.0} & 70.0 & 61.0 & \textbf{75.0} & 45.0 \\
place\_shoe                & 64.0 & 24.0 & \textbf{84.0} & \textbf{84.0} & 78.0 & \textbf{84.0} & 63.0 & 58.0 & 75.0 & 76.0 \\
press\_stapler             & 42.0 & 32.0 & \textbf{86.0} & \textbf{71.0} & 82.0 & 62.0 & 76.0 & 70.0 & 83.0 & 62.0 \\
put\_bottles\_dustbin      & 14.0 & 2.0 & 25.0 & 20.0 & 42.0 & 32.0 & 41.0 & 19.0 & \textbf{50.0} & \textbf{39.0} \\
put\_object\_cabinet       & 22.0 & 0.0 & 41.0 & 20.0 & 30.0 & 19.0 & 39.0 & \textbf{36.0} & \textbf{43.0} & 29.0 \\
rotate\_qrcode             & 50.0 & 2.0 & 31.0 & 17.0 & 54.0 & 16.0 & 54.0 & 36.0 & \textbf{69.0} & \textbf{37.0} \\
scan\_object               & 62.0 & 38.0 & 62.0 & 37.0 & \textbf{66.0} & \textbf{46.0} & \textbf{66.0} & \textbf{46.0} & 56.0 & 33.0 \\
shake\_bottle              & 96.0 & 72.0 & 91.0 & 75.0 & 88.0 & \textbf{89.0} & \textbf{99.0} & 75.0 & 96.0 & 87.0 \\
shake\_bottle\_horizontally & 98.0 & 74.0 & 98.0 & 81.0 & 84.0 & 87.0 & \textbf{99.0} & 84.0 & 97.0 & \textbf{92.0} \\
stack\_blocks\_three       & \textbf{68.0} & 16.0 & 14.0 & 23.0 & 14.0 & \textbf{28.0} & 10.0 & 13.0 & 28.0 & 23.0 \\
stack\_blocks\_two         & 80.0 & 56.0 & 75.0 & 70.0 & 82.0 & \textbf{75.0} & 77.0 & 56.0 & \textbf{95.0} & 61.0 \\
stack\_bowls\_three        & 38.0 & 36.0 & \textbf{50.0} & 41.0 & 40.0 & 43.0 & 48.0 & \textbf{50.0} & 47.0 & 45.0 \\
stack\_bowls\_two          & \textbf{82.0} & 44.0 & 76.0 & \textbf{78.0} & 66.0 & 73.0 & 73.0 & 67.0 & 72.0 & 72.0 \\
stamp\_seal                & 32.0 & 2.0 & 35.0 & \textbf{42.0} & \textbf{46.0} & 30.0 & 34.0 & 30.0 & 44.0 & \textbf{42.0} \\
turn\_switch               & \textbf{56.0} & 40.0 & 55.0 & 42.0 & 40.0 & 42.0 & 53.0 & 41.0 & 50.0 & \textbf{48.0} \\
\midrule
\textbf{Average}           & 54.9 & 27.7 & 62.2 & 50.9 & 61.4 & 51.0 & 64.1 & 49.2 & \textbf{68.1} & \textbf{53.5} \\
\bottomrule
\end{tabular}
\end{table*}

\end{document}